\documentclass[letterpaper, 10 pt, conference]{ieeeconf}  % Comment this line out if you need a4paper

\IEEEoverridecommandlockouts                              % This command is only needed if 
\usepackage{times}
\usepackage{latexsym}
\usepackage{graphicx}
\usepackage{amsmath}
\usepackage{amssymb}
\usepackage{amsfonts}
\usepackage{booktabs}
\usepackage{bbm}
\usepackage{multicol}
\usepackage{multirow}
\usepackage{algorithm}
\usepackage{algpseudocode}
\usepackage{hyperref}
\usepackage[dvipsnames]{xcolor}
\usepackage{float}
\usepackage{wrapfig}
\usepackage[export]{adjustbox}
\definecolor{darkgreen}{RGB}{0,128,0}
\definecolor{cerulean}{rgb}{0.0, 0.48, 0.65}
\usepackage{caption}
\usepackage{subcaption}
\let\labelindent\relax
\usepackage{enumitem}

\usepackage[backend=biber,
            hyperref=true,
            url=false,
            isbn=false,
            doi=false,
            backref=false,
            style=ieee,
            citestyle=numeric-comp,
            sorting=nyt,%none
            block=none]{biblatex}
\usepackage[font=footnotesize]{caption}
\begin{document}

\title{\LARGE \bf
Model-Based Runtime Monitoring with Interactive Imitation Learning 
}
\author{Huihan Liu$^{1}$, Shivin Dass$^{1}$, Roberto Mart{\'i}n-Mart{\'i}n$^{1}$, Yuke Zhu$^{1}$ % <-this % stops a space
\thanks{$^{1}$\fontsize{7}{8}\selectfont The University of Texas at Austin. Correspondence: \url{huihanl@utexas.edu}}%
}

\maketitle
\thispagestyle{empty}
\pagestyle{empty}

\begin{abstract}
Robot learning methods have recently made great strides, but generalization and robustness challenges still hinder their widespread deployment. 
Failing to detect and address potential failures renders state-of-the-art learning systems not combat-ready for high-stakes tasks. 
Recent advances in interactive imitation learning have presented a promising framework for human-robot teaming, enabling the robots to operate safely and continually improve their performances over long-term deployments. Nonetheless, existing methods typically require constant human supervision and preemptive feedback, limiting their practicality in realistic domains. This work aims to endow a robot with the ability to monitor and detect errors during task execution. We introduce a model-based runtime monitoring algorithm that learns from deployment data to detect system anomalies and anticipate failures. Unlike prior work that cannot foresee future failures or requires failure experiences for training, our method learns a latent-space dynamics model and a failure classifier, enabling our method to simulate future action outcomes and detect out-of-distribution and high-risk states preemptively. We train our method within an interactive imitation learning framework, where it continually updates the model from the experiences of the human-robot team collected using trustworthy deployments. Consequently, our method reduces the human workload needed over time while ensuring reliable task execution. Our method outperforms the baselines across system-level and unit-test metrics, with 23\% and 40\% higher success rates in simulation and on physical hardware, respectively.
More information at \textcolor{cerulean}{\url{https://ut-austin-rpl.github.io/sirius-runtime-monitor/}}

\end{abstract}

\section{Introduction}
\label{s:intro}

We have witnessed tremendous progress in learning-based robotics systems in recent years \cite{andrychowicz2020learning, kalashnikov2018qt, brohan2022rt1}. Despite exciting showcases in research settings, they continue to struggle with generalization and reliability issues for widespread deployment. To achieve continual model improvement in trustworthy deployments, a burgeoning body of work~\cite{liu2022robot, mandlekar2020human} has explored the shift from the conventional ``train-then-deploy'' paradigm to ``learning on the job'' with human-robot teams. Particularly, interactive imitation learning (IIL)~\cite{celemin2022interactive} has advocated a framework where the human supervises the robot's execution and performs interventions to handle difficult situations, and the robot continually learns from deployment data. Existing IIL approaches typically require humans to continuously monitor the system and provide imminent feedback, incurring prohibitive human workloads in realistic applications. A critical step toward making these methods practical is incorporating a \emph{runtime monitoring} mechanism, allowing the robots to self-monitor and predict errors during task execution \cite{rm2021}.

\begin{figure}[t]
    \centering
    \includegraphics[width=1.0\linewidth]{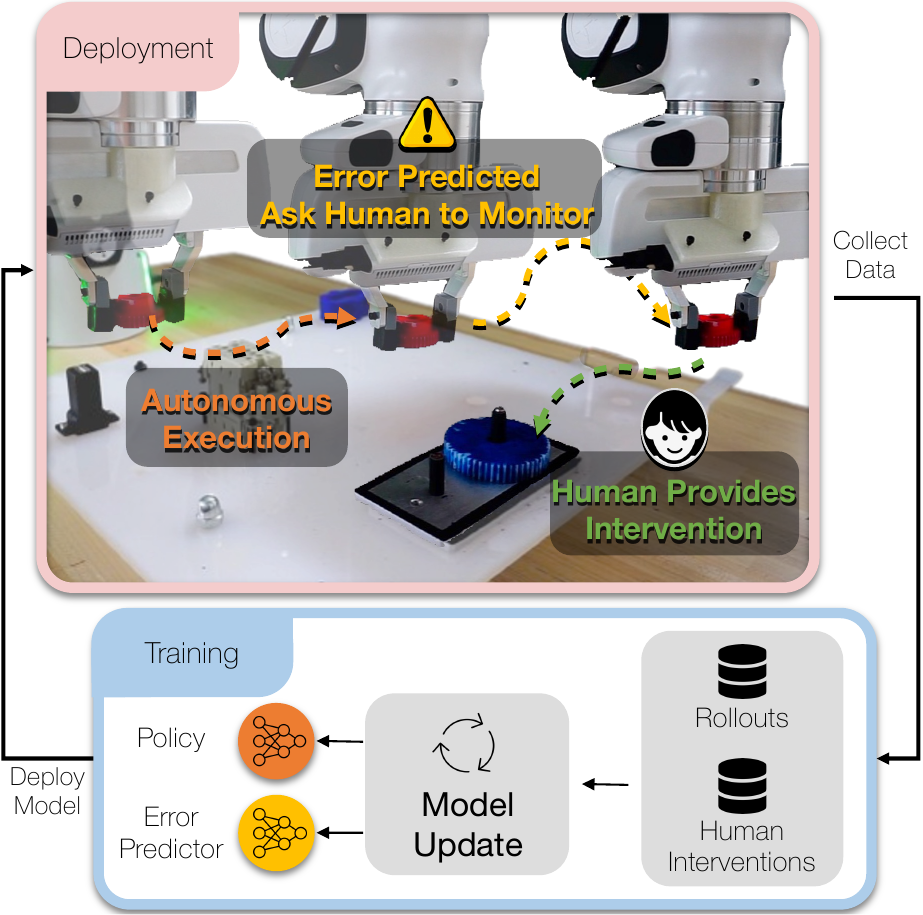}
    \caption{\textbf{Overview.} We introduce a model-based runtime monitoring algorithm that continuously learns to predict errors from deployment data. We integrate this runtime monitoring algorithm into an interactive imitation learning framework to ensure trustworthy long-term deployment.}
    \label{fig:system_workflow}
\end{figure}

Runtime monitoring mechanisms have been investigated in two main directions: unsupervised \emph{out-of-distribution (OOD) detection} and supervised \emph{failure detection}. Unsupervised out-of-distribution (OOD) detection methods~\cite{Richter_OOD, menda2019ensembledagger, momart_wong22a, dass2022pato} use the data distributions as a proxy to estimate errors. They often struggle to faithfully capture the error patterns and may overlook critical situations \cite{ood_survey, DBLP:journals/corr/abs-2003-02977}. More critically, these methods determine whether a failure occurs based on the current system states but cannot identify future failures before they become unavoidable. Supervised approaches train failure detection models from actual failure experiences. Failure detection can be done by classifying normal and failure examples \cite{Gokmen2023AskingFH, xie2022ask} or estimating risk Q-values to predict future failures \cite{hoque2021thriftydagger, hoque2022fleetdagger}. These approaches are more accurate in capturing error patterns and preempting failures. Nonetheless, they heavily rely on explicit failures encountered in the past as supervisory signals. In deployed systems, failures can be catastrophic, and acquiring such failure data hinders the system's safety and reliability.

To this end, we introduce a runtime monitoring algorithm as an extension to Sirius \cite{liu2022robot}, our prior work that improves robot autonomy over long-term deployments with humans in the loop. The runtime monitoring algorithm proactively detects system anomalies and preemptive failures in the interactive imitation learning framework of Sirius (see Fig.~\ref{fig:system_workflow}). Our design follows two conceptual ideas: First, it adopts a \emph{model-based} approach, which trains a predictive model of the environment dynamics for failure prediction. 
The dynamics model can simulate future policy rollouts and predict upcoming failures. 
Second, it uses an \emph{intervention-informed} approach to train its model components over long-term deployments. The algorithm harnesses the inherent structure of human interventions to continually learn an error predictor without having to encounter explicit failures, thus ensuring trustworthy task execution.

Our method performs model-based runtime monitoring with two learnable components: a \textit{dynamics model} and a \textit{failure classifier}. We first construct a latent space, where image observations are encoded into feature vectors as the latent states. We train a dynamics model that predicts the next latent state conditioned on the current observation and the action. We also train a policy from the same latent space. The latent state space shared between the dynamics model and the policy allows our method to simulate counterfactual trajectories and predict different action outcomes \cite{wu2023daydreamer}. We also train a failure classifier that predicts whether a future state leads to failure. With these two components, an error is identified by out-of-distribution (OOD) detection with the dynamics model and failure detection with the dynamics model and the failure classifier. Contrary to prior work \cite{Richter_OOD, hoque2021thriftydagger} that uses isolated OOD and failure detection systems, we find it effective to unify them in a single model, enhancing the data efficiency and overall performance of our system.

Prior learning-based error predictors \cite{gokmen2023asking, xie2022ask, hoque2021thriftydagger} typically require explicit failures to form training datasets, which are infeasible to collect during safe deployments. Instead, we leverage human interventions to inform the error predictor of the moments when humans perceive the system is at risk of failure. 
Human interventions offer implicit cues of their judgment on whether the robot's actions may be incorrect, undesirable, or close to failing \cite{liu2022robot, hoque2021thriftydagger, mandlekar2020human}. We train the failure classifier with human interventions and robot rollouts collected from deployments and continuously update it over time. As a result, the failure classifier can dynamically adapt to the policy's changing behaviors.

We tested our method on two simulated and two real-world tasks with a human-in-the-loop learning and deployment framework \cite{liu2022robot,hoque2021thriftydagger}. 
Our method achieves on average 32\% higher success rates and 16\% more efficient use of human workload utilization than state-of-the-art interactive imitation learning baselines.
In summary, our main contributions are as follows:

\begin{itemize}[leftmargin=*]
    \item We introduce a model-based runtime monitoring algorithm that proactively solicits human help based on future error preemption;
    \item We develop an effective learning method to jointly and continually update our error predictor and policy during human-in-the-loop deployments;
    \item We systematically evaluate our method against baselines in simulation and on physical hardware and demonstrate its effectiveness at the system level and in unit tests.
\end{itemize}

\section{Related Work}

\textbf{Interactive Imitation Learning.} 
In interactive imitation learning \cite{Argall2009ASO}, robots receive human feedback during task execution, allowing for continuous improvements of the policy performances~\cite{celemin2022interactive}. The human involvement in the learning loop has two ways: 1) \textit{human-gated}, where the human constantly supervises the robot and decides when to provide feedback \cite{ross2011reduction, kelly2019hg, liu2022robot}; or 2) \textit{robot-gated}, where the robot actively solicits human feedback~\cite{menda2019ensembledagger, hoque2021thriftydagger, hoque2023fleet}. The robot-gated decision is typically based on heuristic metrics such as ensemble variance \cite{menda2019ensembledagger}, risk \cite{hoque2021thriftydagger}, or task uncertainty \cite{dass2022pato}. While prior robot-gated works have studied efficiently sourcing human feedback for learning, this work investigates automated error prediction with the goal of minimizing runtime errors and ensuring trustworthy deployment.

\textbf{Runtime Monitoring in Robot Learning.} Runtime monitoring and error prediction methods have received considerable interest in robotics \cite{Hsu2023TheSF, yel_runtime, SinhaSchmerlingEtAl2023}.
For robot learning methods, in particular, there have mainly been two bodies of work: out-of-distribution (OOD) detection \cite{ood_survey, sinha2022systemlevel}, and failure detection \cite{diryag2014neural}. Unsupervised out-of-distribution (OOD) detection methods \cite{Richter_OOD, menda2019ensembledagger, momart_wong22a, dass2022pato} use the degree of OOD as an approximate proxy for error prediction. The supervised failure detection approach performs binary classification of normal/failure states \cite{Gokmen2023AskingFH, xie2022ask} with a classifier trained on success and failure examples. A recent line of work~\cite{hoque2021thriftydagger, hoque2022fleetdagger} learns a risk Q function from agent rollouts to identify states that are likely to lead to failures using techniques developed in safe reinforcement learning (RL)~\cite{Brunke2021SafeLI, thananjeyan2021recovery, srinivasan2020learning}. 

\textbf{Model-based Learning Approach.} Learned dynamics models have been adopted in reinforcement learning (RL) \cite{janner2019trust, hafner2019dream, wu2023daydreamer, hafner2023mastering, hansen2022modem}, imitation learning \cite{demoss2023ditto, hu2022modelbased}, planning \cite{Danesh2022LEADERLA, hansen2022temporal, argenson2020model, schubert2023generalist}. They have also demonstrated success in different robotics applications \cite{wu2023daydreamer, mendonca2023structured, mendonca2023alan, shi2022skill}. Prior work uses dynamics learning for safe RL to learn a safe policy given known safety violation criteria \cite{thomas2022safe}. In the same vein, we harness the predictive ability of model-based approaches for runtime monitoring. Concretely, we use the dynamics model to predict the outcomes of future policy actions in a latent space \cite{successor_features}, allowing us to identify risky states \emph{before} they occur and ask for timely human supervision.

\section{Model-based Runtime Monitoring}

We start by describing our problem formulation in Section~\ref{subsec:method_problem_formulation}. We then discuss our algorithmic components in Section~\ref{subsec:method_dyn_and_policy}, and how the individual modules (Section~\ref{subsec:method_runtime_op_comp}) and the overall system (Section~\ref{subsec:method_runtime_op_sys}) are used for runtime monitoring in practice.

\subsection{Problem Formulation}
\label{subsec:method_problem_formulation}

We formulate a robot manipulation task as a discrete-time Markov Decision Process~(MDP), $\mathcal{M}=~(\mathcal{S}, \mathcal{A}, \mathcal{P}, \mathcal{R}, \gamma)$, with continuous states $s\in\mathcal{S}$, continuous actions $a\in\mathcal{A}$, unknown transition dynamics $\mathcal{P}(.|s,a)$, reward function $\mathcal{R}(s,a,s')$ and discount factor $\gamma\in[0,1)$. We aim to learn a task-oriented robot policy $\pi_r:\mathcal{S}\rightarrow\mathcal{A}$, that maximizes the return $\mathbb{E}[\sum_{t=1}^T\mathcal{R}(s_t, a_t, s_{t+1})]$ while minimizing the number of failures during its deployment. 

We consider a human-in-the-loop learning and deployment framework, where a robot performs tasks with humans available to provide corrective feedback in the form of interventions. In contrast to prior work~\cite{liu2022robot, mandlekar2020human, hgdagger2019} where the human continuously monitors the system and provides feedback whenever necessary, our work focuses on developing a runtime monitoring mechanism that actively solicits human feedback only when an error is detected. 

\begin{figure}[t]
    \centering
    \includegraphics[width=0.9\linewidth]{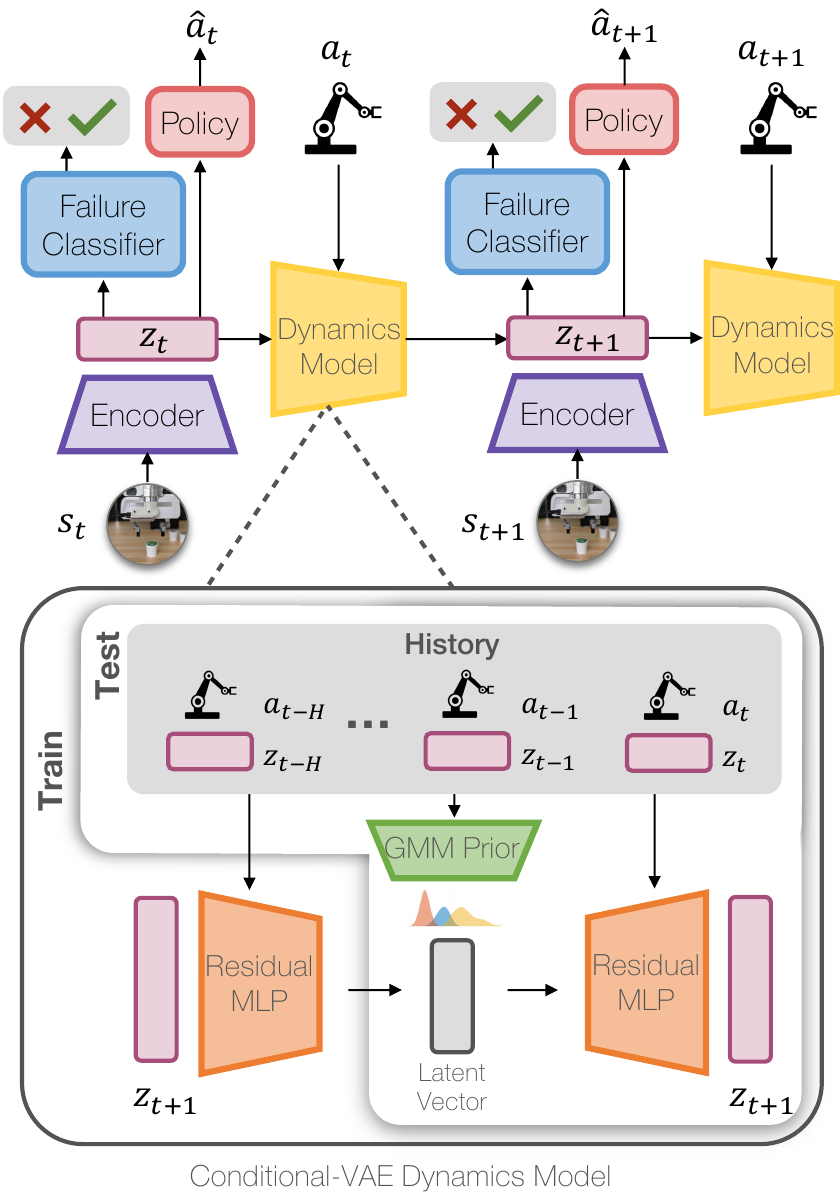}
    \caption{\textbf{Model Architecture.} We train a dynamics model, implemented as a conditional Variational Autoencoder (cVAE), to predict the next latent state given the current state and action. We also train a policy and a failure classifier head based on the latent state. The dynamics model and policy are trained from the experiences collected from task execution. The failure classifier uses the human intervention states to infer failure states.}
    \label{fig:architecture}
\end{figure}

\subsection{Model-based Method Design}
\label{subsec:method_dyn_and_policy}
Here we discuss our model-based approach to runtime monitoring. As shown in Figure~\ref{fig:architecture}, our approach comprises of four main components denoted as $\mathcal{W} = (E_\gamma, T_\psi, \pi_{\theta}, C_\lambda)$. The observation encoder $E_\gamma$ translates raw image and robot proprioceptive state observations into latent state embeddings. These embeddings form a shared latent space wherein the dynamics model $T_\psi$, the policy $\pi_{\theta}$, and the failure classifier $C_\lambda$ all operate. 

\textbf{Dynamics Model and Policy.} The observation encoder $E_\gamma$ first encodes the raw observation into a latent state space representation. The dynamics model $T_\psi$ predicts future latent state rather than future raw observations, enabling fast sampling of futures as noted in prior work~\cite{wu2023daydreamer, hafner2019dream, hansen2022modem}. Given the input sequence of past latent states' embeddings and the current action, the model outputs the next latent state. The dynamics model uses a conditional Variational Autoencoder~(cVAE). The stochastic latent space of cVAE supports the sampling of multiple future predictions, helping to model future outcomes accurately. 

The policy $\pi_{\theta}$ is jointly trained with the dynamics model and shares the same latent embedding space so as to enable policy rollouts in the latent space. The overall training objective combines the maximum likelihood behavior cloning~(BC) objective with the VAE loss terms, which include Kullback-Leibler~(KL) divergence and reconstruction loss of the next state latent embedding: 
\begin{equation}
\begin{aligned}
L_W = \underset{ \substack{(s, a, s') \sim \mathcal{D} \\ q(z|s', s, a)}}{\mathbb{E}} &-\log \pi_{\theta}(a \mid s)
- \log p(s'|s, a, z)\\
&+ D_{KL}(q(z|s', s, a)||p(z|s, a))
\end{aligned}
\end{equation}

\textbf{Failure Classifier.} After jointly training the policy and dynamics model, we train a failure classifier $C_\lambda$ on the frozen latent space. 
Our failure labels come from moments in past interaction data when human intervenes. Based on prior work~\cite{liu2022robot}, we observe that pre-intervention points, \emph{i.e.}, the states right before each human intervention, reflect the human's real-time judgment of undesirable or ``failure'' states. Therefore, these labels act as indicators of human risk assessment and can be used for training the failure classifier.

One particular design choice is that rather than a binary classification of failure versus normal states, we adopt a three-class classification of failure state, normal rollout, and intervention state. This is because we observe that human interventions and system failures often occur in similar states. Therefore, separating these two categories provides a more informative learning signal and  
improves the failure classifier accuracy.
We train this classifier with a cross-entropy loss objective $L_{\text{{C}}} = -\sum_{i=1}^{n} y_i \log(\hat{y}_i)$ with balanced sampling.

\vspace{0.5mm}
\textbf{Implementation Details.} We implement our behavior cloning policy with a backbone of BC-RNN using ResNet encoders~\cite{he2015deep}. We implement the cVAE dynamics model with GMM latent vector~\cite{robomimic} and residual MLP~\cite{wang2020critic} encoder and decoder~(see Figure~\ref{fig:architecture} bottom), making both the latent vector and decoder conditioned on the state and action input. We train the failure classifier with an LSTM, which takes in the history of states and actions and outputs the probability of failure.

\subsection{Runtime Monitor in Operation: Modules}
\label{subsec:method_runtime_op_comp}

During deployment, the system draws samples from the cVAE dynamics model to construct $N$ future policy rollouts in the latent space. This process allows us to predict \emph{l} steps ahead by iteratively running the dynamics model and the policy. 
Since the cVAE model has a stochastic latent space to draw samples from, we can generate multiple predictions of future states. We evaluate each future independently and then average the error predictor results. 
Each predicted future state undergoes a two-level evaluation, which assesses two aspects of the system:

\textbf{Out-of-Distribution (OOD) Detection.} 
We evaluate whether a state is OOD by calculating the nearest neighbor distance to the latent embeddings (obtained from $E_\gamma$) of the in-distribution data, \textit{i.e.}, expert demonstrations collected by humans. We measure the mean distance of multiple imaginary futures and classify a state as OOD if the nearest neighbor distance is above a threshold $\alpha$. 

\textbf{Failure Detection.} If the state is not classified as OOD, Our method anticipates whether there will be a future failure. 
We employ the failure classifier $C_\lambda$ which classifies a state as a failure when the 
averaged failure classification result
across multiple future rollouts exceeds a threshold $\beta$, indicating a high probability that this state will result in a failure. Human supervision is only solicited if $K$ of the past $P$ states are identified as failures, ensuring temporally consistent predictions and reducing the false positive rates. 

Overall, the model-based future prediction enables effective preemptive future state forecasting, and the complementary roles of OOD detection and failure detection ensure reliable real-time monitoring during operation. The detailed hyperparameter choices are listed in Table~\ref{momo-hp}.

\begin{table}[t]
\vspace{0.2cm}
\centering
\begin{tabular}{l|l|c|c|c}
\toprule
\textbf{Modules} & \textbf{Hyperparameter} & \textbf{Value} \\
    \midrule
Dynamics Model & Number of futures, $N$ & 200 \\
(shared) & Future horizon length, $l$ & 10 \\
    \midrule
OOD Detection & Nearest-neighbor threshold, $\alpha$ & 0.06 \\
    \midrule
Failure Detection & Future \% threshold, $\beta$ & 0.6 \\
 & Number of failures in history, $K$ & 5 \\
 & History length, $P$ & 10 \\
    \bottomrule
\end{tabular}
\caption{Hyperparameter choices for our method.}
\label{momo-hp}
\end{table}

\subsection{Runtime Monitoring in Operation: System}
\label{subsec:method_runtime_op_sys}

We now elaborate on the procedure for online data aggregation and training. We initialize a base model
$\mathcal{W}^0=~(E_\gamma^0, \pi_{\theta}^0, T_\psi^0, C_\lambda^0)$ using the initial human demonstration dataset $\mathcal{D}^0$. In every subsequent round $r$ of deployment, the robot executes the policy $\pi_{\theta}^r$, while our error predictor performs runtime monitoring. If our method detects OOD or failure, it requests human monitoring. While monitoring, the humans may choose to \emph{actively intervene} if they see undesirable states from their judgment. We denote the points of \emph{active intervention} as the intervention label $h_{t} \in \{0, 1\}$, an indicator variable that is 1 if the human supervisor actively intervenes at that step. This is what is used to train the failure classifier $C_\lambda^r$. 
Each of the transitions $(s_t, a_t, s_{t+1}, h_{t})$ are then stored in a new dataset $\mathcal{D}^r$. For the next round $r+1$, we retrain our models on the combined dataset $\bigcup_{i=0}^{r}D^i$ to obtain $\mathcal{W}^{r+1}$.
\section{Experiments}

In our experiments, we seek to answer the following questions: 1) How effective is our method at ensuring trustworthy system performance? 2) Can our method utilize the human supervision input effectively? 3) Quantitatively, how accurate is our method at predicting errors?

To answer these questions, we conduct two sets of experiments 1) to evaluate the overall \textit{system-level performance}; 2) to \textit{unit test} the performance of error predictors, each of which we elaborate on in the subsequent sections.

\begin{table}[t]
\vspace{0.2cm}
\centering
\begin{tabular}{p{1.75cm}|l|c|c|c|c}
\toprule
\textbf{Tasks} & \textbf{Methods} & \textbf{Round 1} & \textbf{Round 2} & \textbf{Round 3} \\
\midrule
Nut Assembly & \textbf{Ours} & -- & \textbf{98.5} & \textbf{100.0} \\
       & ThriftyDAgger & 86.6 & 90.0 & 88.1 \\
       & PATO & 82.0 & 80.9 & 87.5 \\
       & MoMaRT & 56.0 & 60.0 & 70.0 \\
\midrule
Threading & \textbf{Ours} & -- & \textbf{93.2} & \textbf{98.7} \\
          & ThriftyDAgger & 79.5 & 75.6 & 77.5 \\
          & PATO & 32.0 & 82.9 & 78.1 \\
          & MoMaRT & 34.0 & 76.0 & 80.0 \\
\midrule
Coffee Pod & \textbf{Ours} & -- & \textbf{83.5} & \textbf{92.5} \\
Packing (real) & ThriftyDAgger & 23.1 & 50.9 & 57.5 \\
\midrule
Gear Assembly & \textbf{Ours} & -- & \textbf{80.0} & \textbf{82.9} \\
(real) & ThriftyDAgger & 44.0 & 62.0 & 70.0 \\
\bottomrule
\end{tabular}
\vspace{5pt}
\caption{\textbf{Combined Policy Performance (in Success Rate).} Our method consistently outperforms the baseline over the rounds. Note that the Round 1 results of Ours are N/A as it uses full human monitoring for warm-start.}
\label{table:hitl_results_policy}
\end{table}

\begin{figure*}[t]
\vspace{0.15cm}
    \centering
    \includegraphics[width=1\linewidth]{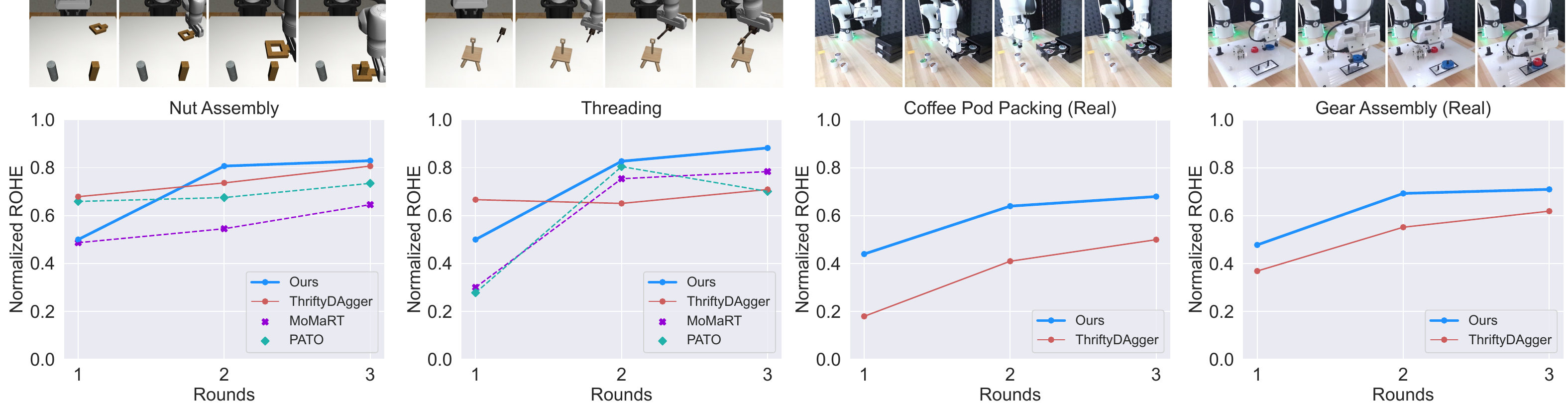}
    \caption{Normalized ROHE curves over three rounds of iterative deployment for tasks Square Nut Assembly~(left), Threading~(left-middle), Coffee Pod Packing~(right-middle), and Gear Assembly~(right). Our method generally has lower ROHE in the first round due to the higher human engagement initially; the ROHE becomes better in later rounds as our method becomes more effective at identifying important errors during deployment.}
    \label{fig:hitl_results}
\end{figure*}

\subsection{System-Level Performance}
\label{subsec:system_performance}

We evaluate the performance of a human-in-the-loop system in an iterative deployment loop. The robot is initially bootstrapped from a BC policy and continues to execute a task with runtime monitoring. The deployment data are aggregated for training the policy to be deployed next round. We run the deployment loop over three repeated rounds, following prior setup \cite{liu2022robot, mandlekar2020human, hgdagger2019}. Each round consists of 1)~data aggregation with the trained policy and human interventions facilitated by the learned error predictors and 2)~training the model over all previously collected data, as described in Section~\ref{subsec:method_runtime_op_sys}. To ensure a fair comparison in the system-level evaluations, we maintain a consistent number of interventions across our approach and the baseline methods. 

\subsubsection{Evaluation Protocol}
\label{subsubsec:eval_protocal}

To benchmark the system-level performance of the runtime monitoring system, we would like to evaluate two aspects: 

\begin{itemize}[leftmargin=*]
    \item \textbf{Collaborative policy performance} measures the overall performance of the system when actively asking for human supervision using the error predictor. It assesses how reliable the runtime monitoring is in guarding the system from failures and facilitating successful task completions. We measure this in the overall task success rate.
    \item  \textbf{Optimized use of human input} measures how the system leverages human input to maximize task outcomes in a fixed time frame. Intuitively, we want the system to engage human assistance precisely when it is most necessary. To quantify this, we adapt the Return On Human Effort (ROHE) from \cite{hoque2023fleet} for a single robot case and normalize it to capture success rates,
    \begin{equation}
        \text{Normalized ROHE}=\frac{\mathbb{E}_\tau[\sum_{t=0}^{T_{\tau}}r^\tau_t]}{1+\frac{H}{T}}
    \end{equation}
    where $H$ is the total number of human interventions across trajectories, $T$ is the total number of time steps, and $r_t^\tau$ is the reward obtained from performing trajectory $\tau$. In sparse reward settings, the numerator is the success rate of the policy while the term in the denominator penalizes the model from asking for excessive human interventions.
    
\end{itemize}

\subsubsection{Baselines}
\label{subsubsec:baselines}
For system-level comparison, we compare our method with \textbf{ThriftyDAgger} \cite{hoque2021thriftydagger}, a state-of-the-art interactive imitation learning baseline that queries human correction in continuous policy updates. ThriftyDAgger detects novel states by training an ensemble of policies and measuring the variance between their outputs. The riskiness of a state is estimated by learning a Q-function from the data and policy rollouts. Note that the additional step of policy rollouts is required for collecting failure data to train the Q-function, which could be infeasible in safety critical domains.

We also compare to two other baselines that were originally not designed for continuous policy update: \textbf{MoMaRT} \cite{momart_wong22a} and \textbf{PATO} \cite{dass2022pato}. MoMaRT uses VAE reconstruction error for the input images as an indicator for out-of-distribution states. PATO uses variance from ensemble policy to measure uncertainty similar to \cite{hoque2021thriftydagger}, as well as the variance of future latent state prediction from VAE to measure the uncertainty of the task. We run these two baselines on the deployment data generated by our method for a fair comparison of the two simulation tasks.

\subsubsection{Results}
\label{subsubsec:exp_tasks} 
We perform two contact-rich manipulation tasks, \textbf{Nut~Assembly} and \textbf{Threading}, in the robosuite simulator \cite{zhu2020robosuite}, and \textbf{Coffee~Pod~Packing} and \textbf{Gear Assembly} in the real world. For all tasks, we use a Franka Emika Panda robot arm equipped with a parallel jaw gripper, and the operational space controller (OSC). 

\begin{table}[t]
    \centering
    \resizebox{1 \columnwidth}{!}{%
    \begin{tabular}{l|c|c|c|c|c|c|c}
    \toprule
    Task & Metrics & \textbf{Ours} & ThriftyDAgger & MoMaRT & PATO  \\
    \midrule
    \multirow{2}{*}{Nut Assembly} & IOU $\uparrow$ & \textbf{0.279} & 0.153 & 0.074 & 0.089 \\
    & DCI $\downarrow$ & \textbf{89.1} & 94.7 & 172.7 & 112.1 \\
    \midrule
    \multirow{2}{*}{Threading} & IOU $\uparrow$ & \textbf{0.317} & 0.044 & 0.077 & 0.078 \\
    & DCI $\downarrow$ & \textbf{30.1} & 60.7 & 129.8 & 47.1 \\
    \midrule
    \multirow{2}{*}{Coffee Pod} & IOU $\uparrow$ & \textbf{0.205} & 0.089 & 0.103 & 0.101 \\
    & DCI $\downarrow$ & \textbf{37.7} & 76.5 & 83.6 & 47.9 \\
    \midrule
    \multirow{2}{*}{Gear Assembly} & IOU $\uparrow$ & \textbf{0.255} & 0.197 & 0.151 & 0.120 \\
    & DCI $\downarrow$ & \textbf{23.5} & 32.2 & 52.7 & 27.9 \\
    \bottomrule
    \end{tabular}
    }
    \vspace{5pt}
    \caption{\textbf{Unit testing error predictors.} Our method outperforms other baselines in the two metrics. Better IOU performance indicates higher overlap between detected and human-labeled failures, and lower DCI means that our method's failure events are closer to the true human failure labeling. }
    \label{tab:shadowing}
\end{table}

We evaluate the collaborative policy performance and the Normalized ROHE score defined in Section~\ref{subsubsec:eval_protocal} at each deployment round. Our method outperforms baselines in terms of collaborative policy performance (see Table~\ref{table:hitl_results_policy}), and Normalized~ROHE score (see Figure~\ref{fig:hitl_results}). We attribute the higher performance of our method to its model-based design: the dynamics model enables future latent state prediction that calls for more timely interventions, resulting in better system performance. 
Note that to train our method's failure classifier, we perform the initial round (Round 1) of deployment with full human monitoring to obtain the human intervention labels. This is reflected in a lower Normalized ROHE score in Round~1 as seen in Figure~\ref{fig:hitl_results}. It, in turn, allows our method to model the failure states well in the later rounds, resulting in a higher combined success rate (see Table \ref{table:hitl_results_policy}) and improved ROHE score (see Round 2 and 3 in Figure \ref{fig:hitl_results}).

In contrast, \textbf{ThriftyDAgger} and \textbf{PATO} employ ensemble policies for OOD detection, which reflects the task uncertainty (multimodality) in the original data, \textit{e.g.}, when the policy is taking gripper actions such as grasping and releasing, resulting in many false positives. \textbf{ThriftyDAgger} also uses a risk Q function to measure the probability of failure. We observe that the values of the trained Q function tend to increase as the robot makes task progress, leading to a higher number of false positives at the beginning of a task and false negatives towards the end, for a fixed threshold value. 
Meanwhile, \textbf{MoMaRT} exhibits some known issues of over-generalization presented in reconstruction-based methods \cite{ood_survey, DBLP:journals/corr/abs-2003-02977}. Specifically, for the fine-grained robot manipulation tasks we examine in this work, failure or success in robot manipulation (e.g. when inserting into a tight hole) might only result in small image variations. VAE reconstruction error often fails to capture such minor pixel differences, leading to numerous false negatives.

\subsection{Unit Testing Error Predictors}
\label{subsec:exp_shadowing}
 
For a more systematic study of error predictor performance, we examine how accurately error predictors capture the actual human risk assessment. We unit test the error predictors' accuracy using \emph{active} human monitoring: we deploy a learned policy within an environment in which a human operator fully supervises the system and can intervene whenever an unsafe state is observed 
\cite{liu2022robot, mandlekar2020human}.
This method enables us to obtain the set of failure states as determined by human observers. 
We then apply the learned error predictors to the collected trajectories. For each trajectory, we obtain the human intervention labels as well as the predicted failure labels. We then compare how close the failure predictions of the error predictors are to that of the human operators.
It enables us to quantitatively measure the performance of the learned error predictors independent of the system deployment loop.

We use two metrics to evaluate the error predictors: 

\begin{itemize}[leftmargin=*]
    \item \textbf{IOU (Intersection over Union)}: we calculate the intersection over union between 
    the predicted failure regions of the error predictor and that of a human.
    We define the intersection and union for our 1D case as follows:
    
    - \emph{Intersection}: Number of timesteps that belong to the predicted failure regions of both the human AND the error predictor. 
    
    - \emph{Union}: Number of timesteps that belong to the predicted failure regions of either the human OR the error predictor.
    
    Intuitively, if the error predictor predicts the failures accurately, the number of timesteps belonging to the intersection would be larger, leading to a higher IOU.

    \item \textbf{DCI (Distance to Closest Intervention)}: for each 
    human intervention, we measure the distance to the closest predicted failure. For each predicted failure, we measure the distance to 
    the human intervention. Finally, we take their average. DCI penalizes predicted failures that are temporally further away from true interventions. 
\end{itemize}

The results in Table~\ref{tab:shadowing} show that our method can accurately predict the intervention points by humans and perform better than the baselines on all tasks. This is mostly attributed to our method's intervention-informed approach. Unlike other methods that uses metrics unrelated to the risk assessment by humans, our method incorporates human intervention data as part of the learning process. By treating human interventions as supervisory signals, our method 
adaptively learns from the decision-making process of human operators that other metrics struggle to capture.
It enables our method to identify potential failures more aligned with human risk assessment than other methods. 

\subsection{Ablation Study}

\begin{figure}[tbp]
    \centering
    \includegraphics[width=1\columnwidth]{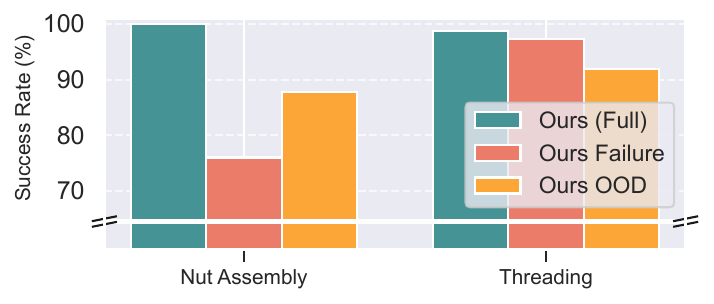}
    \caption{\textbf{Ablation Study.} Ours~(Full) achieves a higher overall performance than Ours Failure and Ours OOD. This result indicates that both failure detection and OOD detection modules complement each other and contribute to the overall success of our system.}
    \label{fig:ablation_results}
    \vspace{10pt}
\end{figure}

\begin{figure}[tbp]
    \centering
    \includegraphics[width=0.95\columnwidth]{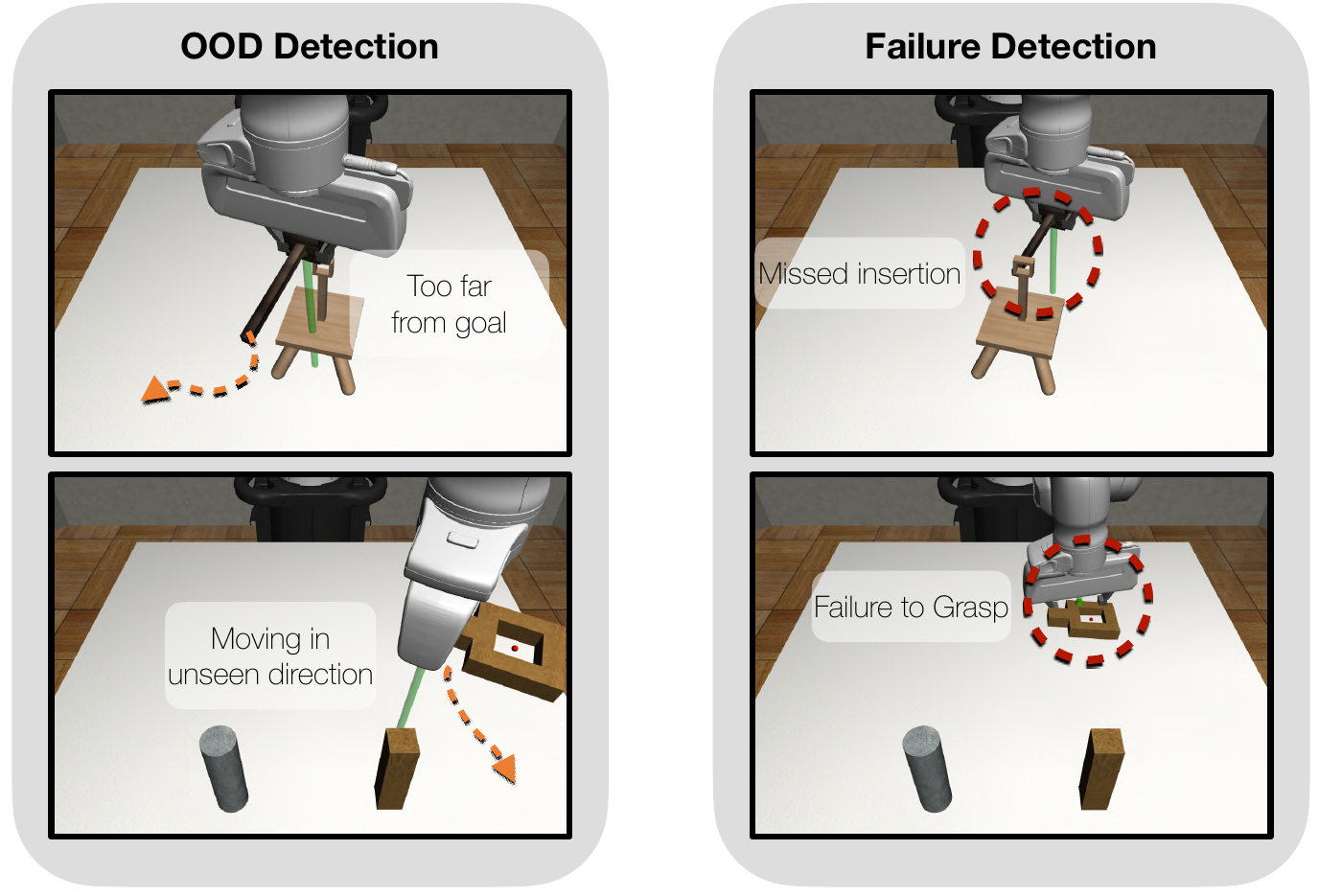}
    \caption{\textbf{Qualitative results from ablation study.} The OOD detection is able to identify unfamiliar states at a coarser level while the failure detection identifies finer-grained failures.}
    \label{fig:qualitative_ablation_results}
    % \vspace{10pt}
\end{figure}

We further evaluate the individual effects of our method's OOD detection and failure detection module on runtime monitoring performance.
We study this in two simulation tasks, Nut Assembly and Threading. We use the Round~3 setting from Table \ref{table:hitl_results_policy} with the same policy and test the  
collaborative policy performance of our method using OOD detection only and failure detection only, denoted as Ours OOD and Ours Failure, respectively. Figure \ref{fig:ablation_results} shows that the combined performance of Ours (Full) outperforms the individual modules, Ours Failure and Ours OOD. We hypothesize that this is because the OOD detection and failure detection modules perform complementary functions. When the robot operates in novel scenarios, OOD detection plays a role in identifying unfamiliar states that may not necessarily be immediate failures but could lead to failures if not addressed. Meanwhile, the failure detection module is capable of identifying subtle changes in critical bottleneck states that OOD detection may overlook. We show example situations where OOD detection and failure detection take effect in Figure~\ref{fig:qualitative_ablation_results}. The two modules work in coordination to identify different kinds of errors, achieving reliable runtime monitoring for continual robot deployment. 

\section{Conclusion and Limitations}
We present a runtime-monitoring algorithm that uses deployment data to predict errors and seek human assistance. We integrate this algorithm into a human-in-the-loop interactive imitation learning system. Results showcase that our method is effective for ensuring trustworthy deployment and facilitating policy learning over long-term deployments. 

Our experiments are conducted by a single human subject per task. The system performance may not faithfully reflect the representative performance of human-robot teams across a broader population. Conducting a large-scale study on human intervention behavior and its impact on our method's ability to capture human-guided failure modes would provide valuable insights. 
Furthermore, our work has focused on quasi-static manipulation tasks, where mistakes are relatively easy to recover from through teleoperation. We are interested in exploring extending our system's applicability to more dynamic environments, diverse robot platforms beyond robot arms, and more complex tasks.
\section*{ACKNOWLEDGMENT}
We thank Ajay Mandlekar for sharing well-designed simulation task environments. We thank Soroush Nasiriany, Yifeng Zhu, Mingyo Seo, Jake Grigsby, Quantao Yang and Yue Wu for providing helpful feedback for this manuscript. We thank Yue Zhao, Bo Liu and Zhenyu Jiang for their fruitful discussions. We acknowledge the support of National Science Foundation (2145283, 2318065), the Office of Naval Research (N00014-22-1-2204), Amazon, and UT Good Systems.

\printbibliography

@inproceedings{ross2011reduction,
  title={A reduction of imitation learning and structured prediction to no-regret online learning},
  author={Ross, St{\'e}phane and Gordon, Geoffrey and Bagnell, Drew},
  booktitle={Proceedings of the fourteenth international conference on artificial intelligence and statistics},
  pages={627--635},
  year={2011},
  organization={JMLR Workshop and Conference Proceedings}
}

@inproceedings{kelly2019hg,
  title={Hg-dagger: Interactive imitation learning with human experts},
  author={Kelly, Michael and Sidrane, Chelsea and Driggs-Campbell, Katherine and Kochenderfer, Mykel J},
  booktitle={2019 International Conference on Robotics and Automation (ICRA)},
  pages={8077--8083},
  year={2019},
  organization={IEEE}
}

@article{mandlekar2020human,
  title={Human-in-the-loop imitation learning using remote teleoperation},
  author={Mandlekar, Ajay and Xu, Danfei and Mart{\'i}n-Mart{\'i}n, Roberto and Zhu, Yuke and Fei-Fei, Li and Savarese, Silvio},
  journal={arXiv preprint arXiv:2012.06733},
  year={2020}
}

@article{hoque2021thriftydagger,
  title={ThriftyDAgger: Budget-aware novelty and risk gating for interactive imitation learning},
  author={Hoque, Ryan and Balakrishna, Ashwin and Novoseller, Ellen and Wilcox, Albert and Brown, Daniel S and Goldberg, Ken},
  journal={arXiv preprint arXiv:2109.08273},
  year={2021}
}

@inproceedings{hoque2023fleet,
  title={Fleet-dagger: Interactive robot fleet learning with scalable human supervision},
  author={Hoque, Ryan and Chen, Lawrence Yunliang and Sharma, Satvik and Dharmarajan, Karthik and Thananjeyan, Brijen and Abbeel, Pieter and Goldberg, Ken},
  booktitle={Conference on Robot Learning},
  pages={368--380},
  year={2023},
  organization={PMLR}
}

@inproceedings{menda2019ensembledagger,
  title={{EnsembleDagger}: A Bayesian approach to safe imitation learning},
  author={Menda, Kunal and Driggs-Campbell, Katherine and Kochenderfer, Mykel J},
  booktitle={2019 IEEE/RSJ International Conference on Intelligent Robots and Systems (IROS)},
  pages={5041--5048},
  year={2019},
  organization={IEEE}
}

@article{dass2022pato,
  title={PATO: Policy Assisted TeleOperation for Scalable Robot Data Collection},
  author={Dass, Shivin and Pertsch, Karl and Zhang, Hejia and Lee, Youngwoon and Lim, Joseph J and Nikolaidis, Stefanos},
  journal={arXiv preprint arXiv:2212.04708},
  year={2022}
}

@article{hafner2019dream,
  title={Dream to control: Learning behaviors by latent imagination},
  author={Hafner, Danijar and Lillicrap, Timothy and Ba, Jimmy and Norouzi, Mohammad},
  journal={arXiv preprint arXiv:1912.01603},
  year={2019}
}

@article{shi2022skill,
  title={Skill-based model-based reinforcement learning},
  author={Shi, Lucy Xiaoyang and Lim, Joseph J and Lee, Youngwoon},
  journal={arXiv preprint arXiv:2207.07560},
  year={2022}
}

@article{hansen2022temporal,
  title={Temporal difference learning for model predictive control},
  author={Hansen, Nicklas and Wang, Xiaolong and Su, Hao},
  journal={arXiv preprint arXiv:2203.04955},
  year={2022}
}

@article{argenson2020model,
  title={Model-based offline planning},
  author={Argenson, Arthur and Dulac-Arnold, Gabriel},
  journal={arXiv preprint arXiv:2008.05556},
  year={2020}
}

@article{demoss2023ditto,
  title={DITTO: Offline Imitation Learning with World Models},
  author={DeMoss, Branton and Duckworth, Paul and Hawes, Nick and Posner, Ingmar},
  journal={arXiv preprint arXiv:2302.03086},
  year={2023}
}

@inproceedings{wu2023daydreamer,
  title={Daydreamer: World models for physical robot learning},
  author={Wu, Philipp and Escontrela, Alejandro and Hafner, Danijar and Abbeel, Pieter and Goldberg, Ken},
  booktitle={Conference on Robot Learning},
  pages={2226--2240},
  year={2023},
  organization={PMLR}
}

@article{hafner2023mastering,
  title={Mastering Diverse Domains through World Models},
  author={Hafner, Danijar and Pasukonis, Jurgis and Ba, Jimmy and Lillicrap, Timothy},
  journal={arXiv preprint arXiv:2301.04104},
  year={2023}
}

@article{hansen2022modem,
  title={MoDem: Accelerating Visual Model-Based Reinforcement Learning with Demonstrations},
  author={Hansen, Nicklas and Lin, Yixin and Su, Hao and Wang, Xiaolong and Kumar, Vikash and Rajeswaran, Aravind},
  journal={arXiv preprint arXiv:2212.05698},
  year={2022}
}

@article{schubert2023generalist,
  title={A Generalist Dynamics Model for Control},
  author={Schubert, Ingmar and Zhang, Jingwei and Bruce, Jake and Bechtle, Sarah and Parisotto, Emilio and Riedmiller, Martin and Springenberg, Jost Tobias and Byravan, Arunkumar and Hasenclever, Leonard and Heess, Nicolas},
  journal={arXiv preprint arXiv:2305.10912},
  year={2023}
}

@article{liu2022robot,
  title={Robot Learning on the Job: Human-in-the-Loop Autonomy and Learning During Deployment},
  author={Liu, Huihan and Nasiriany, Soroush and Zhang, Lance and Bao, Zhiyao and Zhu, Yuke},
  journal={arXiv preprint arXiv:2211.08416},
  year={2022}
}

@misc{hoque2022fleetdagger,
      title={Fleet-DAgger: Interactive Robot Fleet Learning with Scalable Human Supervision}, 
      author={Ryan Hoque and Lawrence Yunliang Chen and Satvik Sharma and Karthik Dharmarajan and Brijen Thananjeyan and Pieter Abbeel and Ken Goldberg},
      year={2022},
      eprint={2206.14349},
      archivePrefix={arXiv},
      primaryClass={cs.RO}
}

@misc{xie2022ask,
      title={When to Ask for Help: Proactive Interventions in Autonomous Reinforcement Learning}, 
      author={Annie Xie and Fahim Tajwar and Archit Sharma and Chelsea Finn},
      year={2022},
      eprint={2210.10765},
      archivePrefix={arXiv},
      primaryClass={cs.LG}
}

@misc{sinha2022systemlevel,
    title={A System-Level View on Out-of-Distribution Data in Robotics},
    author={Rohan Sinha and Apoorva Sharma and Somrita Banerjee and Thomas Lew and Rachel Luo and Spencer M. Richards and Yixiao Sun and Edward Schmerling and Marco Pavone},
    year={2022},
    eprint={2212.14020},
    archivePrefix={arXiv},
    primaryClass={cs.RO}
}

@inproceedings{SinhaSchmerlingEtAl2023,
  author = {Sinha, R. and Schmerling, E. and Pavone, M.},
  title = {Closing the Loop on Runtime Monitors with Fallback-Safe MPC},
  year = {2023},
  booktitle = {{Proc. IEEE Conf. on Decision and Control}},
  url = {/wp-content/papercite-data/pdf/Sinha.Pavone.CDC23.pdf},
  note = {Submitted}
}

@article{rm2021,
   title={Predictive Runtime Monitoring for Mobile Robots using Logic-Based Bayesian Intent Inference},
   url={http://dx.doi.org/10.1109/ICRA48506.2021.9561193},
   DOI={10.1109/icra48506.2021.9561193},
   journal={2021 IEEE International Conference on Robotics and Automation (ICRA)},
   publisher={IEEE},
   author={Yoon, Hansol and Sankaranarayanan, Sriram},
   year={2021},
   month={May} }

@inproceedings{andrychowicz2020learning,
  title={Learning Dexterous In-hand Manipulation},
  author={Marcin Andrychowicz and Bowen Baker and Maciek Chociej and Rafal J{\'o}zefowicz and Bob McGrew and Jakub W. Pachocki and Arthur Petron and Matthias Plappert and Glenn Powell and Alex Ray and Jonas Schneider and Szymon Sidor and Joshua Tobin and Peter Welinder and Lilian Weng and Wojciech Zaremba},
  journal={IJRR},
  year={2018},
  volume={39},
  pages={20 - 3}
}

@inproceedings{kalashnikov2018qt,
  title={{QT-Opt}: Scalable Deep Reinforcement Learning for Vision-based Robotic Manipulation},
  author={Kalashnikov, Dmitry and Irpan, Alex and Pastor, Peter and Ibarz, Julian and Herzog, Alexander and Jang, Eric and Quillen, Deirdre and Holly, Ethan and Kalakrishnan, Mrinal and Vanhoucke, Vincent and others},
  booktitle={CoRL},
  year={2018}
}

@misc{brohan2022rt1,
    title={RT-1: Robotics Transformer for Real-World Control at Scale},
    author={Anthony Brohan and Noah Brown and Justice Carbajal and Yevgen Chebotar and Joseph Dabis and Chelsea Finn and Keerthana Gopalakrishnan and Karol Hausman and Alex Herzog and Jasmine Hsu and Julian Ibarz and Brian Ichter and Alex Irpan and Tomas Jackson and Sally Jesmonth and Nikhil J Joshi and Ryan Julian and Dmitry Kalashnikov and Yuheng Kuang and Isabel Leal and Kuang-Huei Lee and Sergey Levine and Yao Lu and Utsav Malla and Deeksha Manjunath and Igor Mordatch and Ofir Nachum and Carolina Parada and Jodilyn Peralta and Emily Perez and Karl Pertsch and Jornell Quiambao and Kanishka Rao and Michael Ryoo and Grecia Salazar and Pannag Sanketi and Kevin Sayed and Jaspiar Singh and Sumedh Sontakke and Austin Stone and Clayton Tan and Huong Tran and Vincent Vanhoucke and Steve Vega and Quan Vuong and Fei Xia and Ted Xiao and Peng Xu and Sichun Xu and Tianhe Yu and Brianna Zitkovich},
    year={2022},
    eprint={2212.06817},
    archivePrefix={arXiv},
    primaryClass={cs.RO}
}

@InProceedings{momart_wong22a,
  title = 	 {Error-Aware Imitation Learning from Teleoperation Data for Mobile Manipulation},
  author =       {Wong, Josiah and Tung, Albert and Kurenkov, Andrey and Mandlekar, Ajay and Fei-Fei, Li and Savarese, Silvio and Mart{\'i}n-Mart{\'i}n, Roberto},
  booktitle = 	 {Proceedings of the 5th Conference on Robot Learning},
  pages = 	 {1367--1378},
  year = 	 {2022},
  editor = 	 {Faust, Aleksandra and Hsu, David and Neumann, Gerhard},
  volume = 	 {164},
  series = 	 {Proceedings of Machine Learning Research},
  month = 	 {08--11 Nov},
  publisher =    {PMLR},
  url = 	 {https://proceedings.mlr.press/v164/wong22a.html}
}

@article{Gokmen2023AskingFH,
  title={Asking for Help: Failure Prediction in Behavioral Cloning through Value Approximation},
  author={Cem Gokmen and Daniel Ho and Mohi Khansari},
  journal={ArXiv},
  year={2023},
  volume={abs/2302.04334}
}

@article{Richter_OOD,
author = {Richter, Charles and Roy, Nicholas},
year = {2017},
month = {07},
pages = {},
title = {Safe Visual Navigation via Deep Learning and Novelty Detection},
journal={Robotics: Science and Systems},
}

@article{hgdagger2019,
   title={HG-DAgger: Interactive Imitation Learning with Human Experts},
   url={http://dx.doi.org/10.1109/ICRA.2019.8793698},
   DOI={10.1109/icra.2019.8793698},
   journal={2019 International Conference on Robotics and Automation (ICRA)},
   publisher={IEEE},
   author={Kelly, Michael and Sidrane, Chelsea and Driggs-Campbell, Katherine and Kochenderfer, Mykel J.},
   year={2019},
   month={May} }

@misc{he2015deep,
    title={Deep Residual Learning for Image Recognition},
    author={Kaiming He and Xiangyu Zhang and Shaoqing Ren and Jian Sun},
    year={2015},
    eprint={1512.03385},
    archivePrefix={arXiv},
    primaryClass={cs.CV}
}

@misc{robomimic,
    title={What Matters in Learning from Offline Human Demonstrations for Robot Manipulation},
    author={Ajay Mandlekar and Danfei Xu and Josiah Wong and Soroush Nasiriany and Chen Wang and Rohun Kulkarni and Li Fei-Fei and Silvio Savarese and Yuke Zhu and Roberto Martín-Martín},
    year={2021},
    eprint={2108.03298},
    archivePrefix={arXiv},
    primaryClass={cs.RO}
}

@misc{wang2020critic,
    title={Critic Regularized Regression},
    author={Ziyu Wang and Alexander Novikov and Konrad Zolna and Jost Tobias Springenberg and Scott Reed and Bobak Shahriari and Noah Siegel and Josh Merel and Caglar Gulcehre and Nicolas Heess and Nando de Freitas},
    year={2020},
    eprint={2006.15134},
    archivePrefix={arXiv},
    primaryClass={cs.LG}
}

@misc{zhu2020robosuite,
    title={robosuite: A Modular Simulation Framework and Benchmark for Robot Learning},
    author={Yuke Zhu and Josiah Wong and Ajay Mandlekar and Roberto Martín-Martín and Abhishek Joshi and Soroush Nasiriany and Yifeng Zhu},
    year={2020},
    eprint={2009.12293},
    archivePrefix={arXiv},
    primaryClass={cs.RO}
}

@article{diryag2014neural,
  title={Neural networks for prediction of robot failures},
  author={Diryag, Ali and Miti{\'c}, Marko and Miljkovi{\'c}, Zoran},
  journal={Proceedings of the Institution of Mechanical Engineers, Part C: Journal of Mechanical Engineering Science},
  volume={228},
  number={8},
  pages={1444--1458},
  year={2014},
  publisher={SAGE Publications Sage UK: London, England}
}

@article{thananjeyan2021recovery,
  title={Recovery rl: Safe reinforcement learning with learned recovery zones},
  author={Thananjeyan, Brijen and Balakrishna, Ashwin and Nair, Suraj and Luo, Michael and Srinivasan, Krishnan and Hwang, Minho and Gonzalez, Joseph E and Ibarz, Julian and Finn, Chelsea and Goldberg, Ken},
  journal={IEEE Robotics and Automation Letters},
  volume={6},
  number={3},
  pages={4915--4922},
  year={2021},
  publisher={IEEE}
}

@article{srinivasan2020learning,
  title={Learning to be safe: Deep rl with a safety critic},
  author={Srinivasan, Krishnan and Eysenbach, Benjamin and Ha, Sehoon and Tan, Jie and Finn, Chelsea},
  journal={arXiv preprint arXiv:2010.14603},
  year={2020}
}

@article{gokmen2023asking,
  title={Asking for Help: Failure Prediction in Behavioral Cloning through Value Approximation},
  author={Gokmen, Cem and Ho, Daniel and Khansari, Mohi},
  journal={arXiv preprint arXiv:2302.04334},
  year={2023}
}

@article{ood_survey,
  author       = {Mohammadreza Salehi and
                  Hossein Mirzaei and
                  Dan Hendrycks and
                  Yixuan Li and
                  Mohammad Hossein Rohban and
                  Mohammad Sabokrou},
  title        = {A Unified Survey on Anomaly, Novelty, Open-Set, and Out-of-Distribution
                  Detection: Solutions and Future Challenges},
  journal      = {CoRR},
  volume       = {abs/2110.14051},
  year         = {2021},
  url          = {https://arxiv.org/abs/2110.14051},
  eprinttype    = {arXiv},
  eprint       = {2110.14051},
  bibsource    = {dblp computer science bibliography, https://dblp.org}
}

@article{DBLP:journals/corr/abs-2003-02977,
  author       = {Zhisheng Xiao and
                  Qing Yan and
                  Yali Amit},
  title        = {Likelihood Regret: An Out-of-Distribution Detection Score For Variational
                  Auto-encoder},
  journal      = {CoRR},
  volume       = {abs/2003.02977},
  year         = {2020},
  url          = {https://arxiv.org/abs/2003.02977},
  eprinttype    = {arXiv},
  eprint       = {2003.02977},
  bibsource    = {dblp computer science bibliography, https://dblp.org}
}

@misc{janner2019trust,
    title={When to Trust Your Model: Model-Based Policy Optimization},
    author={Michael Janner and Justin Fu and Marvin Zhang and Sergey Levine},
    year={2019},
    eprint={1906.08253},
    archivePrefix={arXiv},
    primaryClass={cs.LG}
}

@misc{thomas2022safe,
    title={Safe Reinforcement Learning by Imagining the Near Future},
    author={Garrett Thomas and Yuping Luo and Tengyu Ma},
    year={2022},
    eprint={2202.07789},
    archivePrefix={arXiv},
    primaryClass={cs.LG}
}

@article{successor_features,
  author       = {Lucas Lehnert and
                  Michael L. Littman},
  title        = {Successor Features Support Model-based and Model-free Reinforcement
                  Learning},
  journal      = {CoRR},
  volume       = {abs/1901.11437},
  year         = {2019},
  url          = {http://arxiv.org/abs/1901.11437},
  eprinttype    = {arXiv},
  eprint       = {1901.11437},
  bibsource    = {dblp computer science bibliography, https://dblp.org}
}

@misc{hu2022modelbased,
      title={Model-Based Imitation Learning for Urban Driving}, 
      author={Anthony Hu and Gianluca Corrado and Nicolas Griffiths and Zak Murez and Corina Gurau and Hudson Yeo and Alex Kendall and Roberto Cipolla and Jamie Shotton},
      year={2022},
      eprint={2210.07729},
      archivePrefix={arXiv},
      primaryClass={cs.CV}
}

@misc{mendonca2023structured,
      title={Structured World Models from Human Videos}, 
      author={Russell Mendonca and Shikhar Bahl and Deepak Pathak},
      year={2023},
      eprint={2308.10901},
      archivePrefix={arXiv},
      primaryClass={cs.RO}
}

@misc{mendonca2023alan,
      title={ALAN: Autonomously Exploring Robotic Agents in the Real World}, 
      author={Russell Mendonca and Shikhar Bahl and Deepak Pathak},
      year={2023},
      eprint={2302.06604},
      archivePrefix={arXiv},
      primaryClass={cs.RO}
}

@article{Danesh2022LEADERLA,
  title={LEADER: Learning Attention over Driving Behaviors for Planning under Uncertainty},
  author={Mohamad H. Danesh and Panpan Cai and David Hsu},
  journal={ArXiv},
  year={2022},
  volume={abs/2209.11422},
  url={https://api.semanticscholar.org/CorpusID:252519500}
}

@inproceedings{Hsu2023TheSF,
  title={The Safety Filter: A Unified View of Safety-Critical Control in Autonomous Systems},
  author={Kai-Chieh Hsu and Haimin Hu and Jaime Fern{\'a}ndez Fisac},
  year={2023},
  url={https://api.semanticscholar.org/CorpusID:261697421}
}

@INPROCEEDINGS{yel_runtime,
  author={Yel, Esen and Bezzo, Nicola},
  booktitle={2019 IEEE/RSJ International Conference on Intelligent Robots and Systems (IROS)}, 
  title={Fast Run-time Monitoring, Replanning, and Recovery for Safe Autonomous System Operations}, 
  year={2019},
  volume={},
  number={},
  pages={1661-1667},
  doi={10.1109/IROS40897.2019.8968498}}

@article{Brunke2021SafeLI,
  title={Safe Learning in Robotics: From Learning-Based Control to Safe Reinforcement Learning},
  author={Lukas Brunke and Melissa Greeff and Adam W. Hall and Zhaocong Yuan and Siqi Zhou and Jacopo Panerati and Angela P. Schoellig},
  journal={ArXiv},
  year={2021},
  volume={abs/2108.06266},
  url={https://api.semanticscholar.org/CorpusID:237048469}
}

@misc{celemin2022interactive,
      title={Interactive Imitation Learning in Robotics: A Survey}, 
      author={Carlos Celemin and Rodrigo Pérez-Dattari and Eugenio Chisari and Giovanni Franzese and Leandro de Souza Rosa and Ravi Prakash and Zlatan Ajanović and Marta Ferraz and Abhinav Valada and Jens Kober},
      year={2022},
      eprint={2211.00600},
      archivePrefix={arXiv},
      primaryClass={cs.RO}
}

@article{Argall2009ASO,
  title={A survey of robot learning from demonstration},
  author={Brenna Argall and S. Chernova and Manuela M. Veloso and Brett Browning},
  journal={Robotics Auton. Syst.},
  year={2009},
  volume={57},
  pages={469-483},
  url={https://api.semanticscholar.org/CorpusID:1045325}
}

\end{document}